# Analytic Robot Self-Model: Learning Screw-Axis Kinematics for Rendering and Model-Based Control

Xu Chen, Ning Tan*

School of Computer Science and Engineering, Sun Yat-sen University, Guangzhou, China

*Corresponding author: tann5@mail.sysu.edu.cn

## Abstract

Visual self-modelling lets robots learn body models from observations, but such models usually encode kinematics implicitly rather than exposing metric joint parameters for computation and control. Moreover, a good fit to images does not by itself establish accurate kinematics. Here we present an explicit analytic robot self-model that requires neither deformation networks nor URDF/CAD templates. Multi-view images and joint readings establish Gaussian appearance and provide motion evidence for rigid link assignments, whereas sparse end-effector poses independently identify metric screw axes. The resulting closed-form forward kinematics and analytic Jacobian can be evaluated beyond the fitting ranges. We validate the model on a simulated seven-axis arm, a simulated twelve-joint quadruped and a real six-axis arm. In simulation, it achieves millimetre-level reaching and pick-and-place, as well as closed-loop quadruped walking. On the real arm, it renders held-out configurations, predicts end-effector positions with a median error of 6.4 mm, compared with 11.7 mm for a calibrated nominal model, and supports reaching with external marker feedback. A controlled comparison shows that image fit alone is not enough: two Panda models differing only in their joint axes are 0.55 dB apart in PSNR, yet their end-effector prediction errors differ 13.7-fold.

## Introduction

Robot planning and control typically rely on geometry and kinematic parameters specified in advance by engineers, whereas self-modelling lets the robot obtain this information from its own sensing and motion, an attractive option when a robot is hard to model precisely, its structure is unknown or its body changes during use [1–5]. To replace hand-specified parameters, however, a self-model must do more than describe how the robot looks in different configurations: it must also predict motion in physical space and expose kinematic quantities that a solver can use directly.

Recent visual self-models learn configuration-dependent body representations from images and joint states, and use them for prediction, planning and closed-loop control [6–10]. They typically do not expose metric joint axes or a composable analytic chain. In parallel, visual articulation reconstruction recovers appearance, parts and explicit joint parameters, mostly for everyday articulated objects [11–16]. ScrewSplat does this from multi-view RGB by optimizing 3D Gaussians and screw axes jointly, but assumes that every movable part articulates with respect to a single static base and does not consider kinematic chains [16]. Two other lines of work also build explicit articulated models, but for different ends. ChainSplat combines a screw-theoretic open chain with Gaussian splatting to approximate deformable linear objects such as ropes and to learn their dynamics, rather than to model robots [17]. Methods based on category priors, code generation or vision–language models produce simulation-ready articulated assets from images or meshes [18–21], but they aim for plausible, usable models, and their joint parameters are not calibrated against physical measurements.

On robot bodies, some methods build renderable representations from templates or known kinematics [22,23]. Others infer parts, topology and joint parameters from metrically scaled point clouds [24]. Still others align a visual reconstruction to the robot frame and recover metric scale, but take the robot pose or its kinematics as given [25,26]. Separately, work on radiance fields has found that geometry and view-dependent radiance can trade off against each other, so that different geometries fit the same images [27,28]. This raises two related questions. Can a robot obtain, from its own observations, a model that both renders and supplies metric analytic kinematics? Can the model's own measures of fit tell us whether that kinematics is accurate?

Here we present an explicit analytic robot self-model based on 3D Gaussian splatting [29], without deformation networks or URDF/CAD templates. We learn a renderable appearance and rigid link assignments from multi-view images paired with joint readings. Metric joint axes identified from relative end-effector motion supply the analytic rigid transforms used for rendering and kinematic computation. Differentiating the chain gives an analytic Jacobian for control. We evaluate the model on a simulated arm, a simulated quadruped and a real arm, and use it in simulated tasks and real-robot reaching. Controlled comparisons reveal an order-of-magnitude difference in end-effector error despite similar rendering quality.

## An explicit analytic self-model unifies appearance and kinematics

The model consists of static 3D Gaussians at a reference configuration, each carrying a rigid link label, and every revolute or prismatic joint is described by a screw axis. Given the joint readings, rigid transforms are composed in order along the joint path from the robot's root to each link, and all Gaussians on that link share the resulting transform. The model therefore renders the robot at new configurations without a per-frame deformation network and directly provides closed-form forward kinematics and an analytic Jacobian (Fig. 1a).

Multi-view images at the reference configuration provide the Gaussian appearance, and single-joint excitation images provide motion evidence for link assignment. When one joint leaves the reference configuration, its downstream subtree moves as a whole while upstream links and other branches stay still. Combining these patterns of motion and rest across joints, we learn each Gaussian's link assignment. Panda's link6 has a nearly axisymmetric cylindrical surface whose appearance changes little when it rotates about its own axis, so we resolve this local ambiguity with a part-mask constraint. The assignment and refinement procedures for each robot are described in Methods.

We apply this representation to a simulated seven-axis Panda, a simulated Laikago with twelve revolute joints on four branches, and a real six-axis JAKA arm (Fig. 2). Serial chains and trees share the same evaluation formula; only the parent–child connectivity of the joints changes. Under their respective evaluation protocols, Panda, Laikago and JAKA reach mean full-image PSNRs of 31.8, 26.9 and 31.0 dB on 4,000, 213 and 31 test images, respectively.

The analytic model can also be evaluated beyond the fitting ranges. Every Panda joint is tested 4°–61° beyond the range covered by the fitting observations, while the displayed configurations reach joint angles up to 210° with intact renderings (Fig. 1b; Supplementary Table 1b). On JAKA, the 31 test images comprise 10 single-joint and 21 multi-joint configurations from one camera station; the multi-joint subset reaches a mean PSNR of 30.37 dB and silhouette IoU of 0.8416 (Fig. 2c; Supplementary Table 2a).

The learned prismatic gripper motion transfers to configurations in which the arm and gripper move together. We fit the Panda gripper translation direction from images in which the arm is fixed and the gripper sweeps five openings, without three-dimensional direction ground truth. On 4,000 unseen frames with jointly varying arm configuration and gripper opening, rendering the observed opening improves the gripper region over rendering it closed, and the gain increases with opening (Supplementary Table 3a).

## Sparse end-effector observations anchor visual kinematics in physical space

Whether the analytic chain predicts the robot's motion accurately in physical space depends on the metric accuracy of the joint axes. We identify these axes from relative end-effector motion under single-joint excitation, without knowing the downstream geometry or the mounting transform of the end-effector marker. When only joint $j$ moves, the end-effector motion relative to the reference depends on that joint's screw axis and angle alone, because the downstream fixed transforms and the marker mounting offset cancel. Each axis therefore follows in closed form from the motion it alone produces (Fig. 3a; Methods Section 4.1). Against simulator kinematic ground truth, the axes on Panda and Laikago have mean direction errors of 0.045° and 0.097°, and mean shortest inter-axis distances of 0.42 mm and 0.51 mm, respectively (Supplementary Table 4a).

Independent end-effector observations substantially improve motion prediction. In the Panda axis-substitution pair, replacing image-supervised axes with metric-anchored axes reduces mean end-effector prediction error from 34.19 mm to 2.49 mm while Gaussians, link assignments, cameras and evaluation configurations are held fixed (Fig. 6b; Methods Section 6.2).

The learned kinematics also predicts motion on the real JAKA arm. Against external marker measurements at 15 configurations, the self-model has a median forward-kinematic position error of 6.4 mm, compared with 11.7 mm for the calibrated nominal baseline, and has lower error at 13 of the 15 configurations (Fig. 5b; Supplementary Table 7).

In the separate simultaneous-calibration study, eight Panda poses are sufficient to reach sub-millimetre held-out error in the tested configurations, whereas four give unstable solutions (Fig. 3b; Supplementary Table 4b). Parameter counting gives a necessary lower bound of six poses for the 34 effective unknowns (Methods Section 4.3). Prediction error increases with simulated position and orientation noise (Fig. 3c; Supplementary Table 4c).

## Analytic kinematics directly supports model-based control

Metric-anchored kinematics improves executed reaching on Panda. Across 200 world-coordinate targets, damped least squares gives a mean after-execution error of 2.280 mm, with every target within 5 mm. By contrast, substituting image-supervised axes raises the mean error to 7.636 mm and worsens all 200 targets (Fig. 4a; Supplementary Table 5a). The advantage carries over to manipulation: in 50 paired pick–carry–place trials, both models satisfy the task tolerances, but mean grasp-point and placement errors are 2.78 and 3.21 mm with the metric-anchored model, against 13.18 and 19.16 mm with the image-supervised model (Fig. 4b; Supplementary Table 5b). Because the gripper opening is part of the model, the controller can also choose it from the object size: an adaptive opening satisfies the grasp geometry in all 108 tested conditions, whereas a fixed opening satisfies none (Supplementary Table 3b; Methods Section 6.3).

On the quadruped, each foot is controlled through its own branch. With the body fixed, the solved configurations reach 200 foot targets with a median error of 1.162 mm. With the body free under gravity and ground contact, the robot completes four crawl cycles and advances about 250 mm with a maximum tilt of 1.1° (Fig. 5a; Supplementary Table 6).

On the real arm, the learned kinematics supports reaching to 15 marker-defined targets displaced 79.2–228.2 mm from the reference centre, with each joint within ±20° of zero. Using their respective inverse-kinematics implementations, the self-model yields executable solutions for 14 targets and the nominal baseline for 12. On the 12 targets completed by both, median open-loop errors are 6.0 and 4.8 mm, respectively. Marker feedback through the self-model's analytic Jacobian reduces its median error from 6.2 to 4.3 mm, and the error decreases in 12 of the 14 executions (Fig. 5b; Supplementary Table 7).

## Internal fit measures do not establish metric accuracy

The preceding experiments evaluate prediction and execution against external physical references. We next ask whether a model's image-fitting quality and inverse-kinematics residual alone can reveal whether its kinematics is accurate.

In the Panda axis-substitution pair, the two models share Gaussians, link assignments, cameras and test configurations and differ only in their joint axes. Their white-background PSNR differs by about 0.55 dB and their silhouette and pixel-error measures are similar, yet their mean end-effector prediction errors are 34.19 mm and 2.49 mm, a 13.7-fold difference (Fig. 6a,b). Similar image quality can therefore conceal markedly different motion-prediction accuracy. Keeping both models unchanged and switching only the evaluation background to black reverses the PSNR ranking, with the image-supervised model scoring about 0.19 dB higher, although neither the kinematics nor the end-effector predictions change. A ranking by rendering score thus cannot stand in for a ranking by physical motion accuracy.

The same distinction appears in inverse kinematics. For the same Panda reaching targets, damped least squares brings both models to a mean in-model residual of about 0.03 mm, so the solver has placed each model's predicted end-effector close to the target. Executing the resulting joint commands, however, gives mean reaching errors of 2.280 mm for the metric-anchored model and 7.636 mm for the image-supervised model (Fig. 6c). That a solver can solve a model precisely does not mean that the model describes the robot precisely. The Laikago comparison goes further: the image-supervised model has the smaller in-model residual yet a foot error nearly an order of magnitude larger, and under the same gait and controller settings it also travels a shorter distance and tilts further, consistently across all 16 tested combinations of friction coefficient and torque limit (Figs. 5a and 6d; Supplementary Table 6b).

To test whether image loss always favours the correct joint axes, we keep the Laikago Gaussian representation and single-joint observations fixed and substitute either the simulator ground-truth axes or the image-supervised axes. Three of the twelve joints, all in the calves, give lower image loss under the image-supervised axes, and for the four hip joints the losses differ by only a few parts in ten thousand. Under these observation conditions, image loss may barely distinguish between axes or may favour an axis that departs from ground truth. A good visual fit or a small inverse-kinematics residual therefore cannot replace evaluating motion accuracy against external end-effector positions and execution results.

## Discussion

We have shown how observations can establish a robot self-model that both renders and computes kinematics. Image-learned Gaussian appearance and rigid link assignments are combined with independently identified metric axes in one explicit representation for rendering, forward kinematics and Jacobian computation. Because the axes can be replaced while appearance, assignments and cameras stay fixed, the same representation also supports controlled comparisons, which let us evaluate end-effector predictions and task execution alongside the images the model produces.

Neither image fit nor in-model residual establishes whether the kinematics is accurate. Rendering metrics measure agreement with images, whereas the in-model residual measures how closely a controller can satisfy a target inside the model. Neither directly tests where the physical robot arrives when the command is executed. A self-model intended for control therefore needs an independent kinematic reference, such as end-effector prediction error, task-space reaching error or joint-axis geometry where a reference model exists. Image fitting and solver convergence remain useful, but they do not replace these physical references.

Rotational symmetry can weaken the motion evidence in images. When Panda's link6 rotates about its own central axis, its approximately uniform cylindrical surface produces little change in silhouette or appearance. Although its Gaussians move with the joint, the images provide little evidence for distinguishing rotation from rest. We therefore constrain the assignment of this local part with a mask, supplying information that motion images leave ambiguous. This example shows that link identification depends not only on joint excitation, but also on whether a part's shape and appearance turn that motion into a distinguishable visual signal.

The appearance representation remains limited by its use of static 3D Gaussians. Surfaces that remain occluded in the reference configuration lack reconstruction observations. Lighting conditions during capture are also encoded in the Gaussian colours. As the robot moves away from the reference configuration, surfaces change orientation relative to the light sources while their colours remain fixed, producing shading inconsistent with the new geometry. Gaussian positions are determined by the analytic kinematic chain independently of colour, so this lighting mismatch affects rendered shading rather than kinematic evaluation. Explicitly separating material properties, illumination and shadows could improve appearance consistency over large joint motions [30]

In the real-robot experiments, encoder readings drive image comparisons, while external markers provide measurements for end-effector prediction, reaching and feedback. Hardware validation thus covers both image reproduction and task-space execution. The measured error combines kinematic model error, marker measurement error and residual coordinate alignment error, and describes the performance of the experimental system as a whole. The real-control evaluation covers 15 targets on one six-axis arm. Further work will examine broader task and configuration ranges, stability during long-term closed-loop operation, and extensions to closed-chain mechanisms and continuum robots.

## Funding

This work was supported by the Open Research Fund of Ningbo Key Laboratory of Spatial Intelligence and Digital Derivative.

# Methods

## 1 · An analytic robot self-model

**Model and observations.** The self-model consists of static 3D Gaussians at a reference configuration, a rigid link label for each Gaussian, a screw axis for each joint and the parent–child connectivity of the joints. The Gaussians, labels and axes are estimated from observations, whereas the connectivity is given. The robot has $n$ joints and its configuration $\mathbf{q} \in \mathbb{R}^n$ is read from the encoders. A calibrated camera $v$ records an image $I_v(\mathbf{q})$ with known intrinsics and extrinsics. At a small number of configurations, we also obtain an externally measured end-effector pose $T_{EE}^{\text{obs}} \in SE(3)$ paired with the corresponding $\mathbf{q}$.

**Posing the model.** The rigid transform of link $l$ at configuration $\mathbf{q}$, measured from the reference configuration, is

$$T_l(\mathbf{q}) = \prod_{j \in \mathcal{A}(l)} \exp\left(\hat{\xi}_j q_j\right). \qquad (1)$$

The product is ordered along the path from the base to link $l$. The set $\mathcal{A}(l)$ is determined by the given connectivity. The matrix representation of the screw axis is

$$\hat{\xi}_j = \begin{bmatrix} [\boldsymbol{\omega}_j]_\times & \\ \mathbf{0}^\top & \end{bmatrix}.$$

For a revolute joint, $\boldsymbol{\omega}_j$ is a unit direction, $\mathbf{c}_j$ is a point on its axis and $\mathbf{v}_j = -\boldsymbol{\omega}_j \times \mathbf{c}_j$. The matrix $[\boldsymbol{\omega}_j]_\times$ represents the cross product. For a prismatic joint, $\boldsymbol{\omega}_j = \mathbf{0}$ and $\mathbf{v}_j$ is the unit translation direction. Revolute coordinates are in radians and prismatic coordinates in metres. This is the product-of-exponentials form of forward kinematics [31,32] All three robots use numerical joint zero as their reference configuration.

For Gaussian $i$, let $\boldsymbol{\mu}_i^0$, $\Sigma_i^0$ and $z_i$ denote its reference centre, covariance and link label. At configuration $\mathbf{q}$,

$$\tilde{\boldsymbol{\mu}}_i(\mathbf{q}) = T_{z_i}(\mathbf{q})\tilde{\boldsymbol{\mu}}_i^0, \qquad \Sigma_i(\mathbf{q}) = R_{z_i}(\mathbf{q})\Sigma_i^0 R_{z_i}(\mathbf{q})^\top,$$

where a tilde denotes a homogeneous point and $R_l$ is the rotation part of $T_l$. Colour and opacity keep their reference values. A tree uses the appropriate ancestor path for each link without changing the evaluation formula.

**Frames.** In simulation, Gaussians, axes, cameras and end-effector references are expressed in the simulator world frame. On JAKA, camera calibration, board observations and end-effector marker measurements establish a fixed registration between model and measurement frames. Registration is estimated from calibration data and held fixed for the corresponding observations and executions. Panda world-coordinate reaching also requires the reference end-effector position $\mathbf{p}_0$ (Section 5.3).

## 2 · Data acquisition

**Images and end-effector observations.** Three kinds of images are used: static multi-view images at the reference configuration, which establish the Gaussian representation; single-joint excitation images, in which only one joint at a time leaves the reference configuration and which provide motion evidence for link assignment and image-supervised axis estimation; and images at multi-joint configurations, which support model refinement and rendering evaluation. Every image is paired with joint readings and camera parameters, and data sizes and configuration ranges are listed in Supplementary Table 1. End-effector poses are provided directly by the simulator in simulation, and on JAKA by an external camera that estimates the pose of a visual marker rigidly attached to the end-effector. The mounting relation remains fixed during acquisition.

Every image also has a robot foreground mask. In simulation, the masks are supplied directly by the simulator, whereas JAKA masks are extracted from RGB images with SAM 3 and cleaned by connected components [33] For Panda's nearly axisymmetric link6, a local part mask is obtained in the same way, with the operator providing a description of the part's appearance. JAKA and Laikago link assignment uses no part-semantic masks.

**Camera and JAKA images.** JAKA images are recorded with the RGB stream of an Intel RealSense D435i at 1920 × 1080 pixels, without depth, and remapped to an 800 × 800 virtual pinhole camera, with the corresponding intrinsics and extrinsics stored for each image. Features on a table-mounted calibration board determine the camera extrinsics, and a square end-effector marker with a 38 mm side provides pose measurements [34] The JAKA multi-pose set contains 202 images from one camera station, split into 146 training, 25 validation and 31 test images; the test split contains 10 single-joint and 21 multi-joint configurations.

**Inputs to the model.** Self-model fitting uses calibrated cameras, joint readings, the known parent–child connectivity, foreground masks, sparse end-effector observations and a local part mask used only for Panda link6. The robot's URDF, CAD geometry and manufacturer-supplied axes are not inputs. In simulation, PyBullet uses the robot description only to generate observations and evaluation references, and contact simulation uses the mass, inertia and contact parameters it provides [35]

## 3 · Gaussian representation and link assignment

Each reference Gaussian has a centre, a covariance parameterized by a scale vector and a rotation quaternion, a colour and an opacity, and the Gaussians are rasterized with 3D Gaussian Splatting [29]. Static reconstruction minimizes pixelwise L1 and structural-similarity losses on multi-view images [36]  Panda and Laikago use a white background, spherical harmonics up to degree 3 and an initial position learning rate of $1.6 \times 10^{-4}$, with 7,000 training steps each. JAKA uses the same losses on white-background RGB images together with foreground-opacity supervision, and the model retained at step 20,000 contains 190,731 Gaussians. Further settings are given in Supplementary Notes 1 and 2.

### 3.1 Estimating initial link assignments

**Moving sets from single-joint excitation.** For each joint $j$, we fit a score $m_{ij} \in [0,1]$ for every Gaussian by comparing rendered and observed excitation images; a score near 1 indicates that the Gaussian moves with the joint and a score near 0 that it stays still. The axes are either fitted jointly with these scores or held at independently identified metric values. Thresholding the scores of all joints gives moving sets, and the ancestry of the chain then determines each link label: a Gaussian on link $l$ should move with the joints in $\mathcal{A}(l)$, which in a serial chain form a consecutive sequence from the base to that link.

**Continuous link weights.** An alternative parameterization gives each Gaussian a vector of link scores $s_i$, converted to weights $w_i = \mathrm{softmax}(s_i/\tau)$. At each observed configuration, the candidate link transforms are blended with dual quaternions under $w_i$ [37] to pose the Gaussians for rendering. Only the assignment scores are optimized, while the axes and the reference Gaussian geometry and colours stay fixed. The loss combines $0.8L_1 + 0.2(1 - \mathrm{SSIM})$, a silhouette term and a spatial-neighbourhood consistency term, and the temperature $\tau$ decreases during training so that the weights concentrate on single links; discrete labels are then $z_i = \mathrm{argmax}_l w_{il}$.

**Local part constraint.** For the axisymmetric Panda link6, we project the reference Gaussians into the SAM 3 part-mask views and aggregate the labels of visible projections. This constraint stays fixed during later refinement, while the other assignments continue to be learned from motion images.

### 3.2 Assignment correction and Gaussian cleanup

**Reprojection voting.** Each candidate link label of a Gaussian is tested by applying the corresponding transform from equation (1) at every training configuration and projecting the Gaussian into the image; the score is the fraction of projections inside the robot foreground silhouette. Voting can use the Gaussian centre alone or also its extent, sampling the centre and two points at $\pm 3\sigma$ on each covariance principal axis (seven points in total). A label is replaced only when the candidate score and its improvement meet the specified thresholds.

**Spatial neighbourhood correction.** Because misassigned Gaussians inside the foreground can escape silhouette voting, we also connect Gaussians by spatial proximity at the reference configuration and identify small same-label connected components. For each Gaussian in such a component, we count the labels of its neighbours outside the component, and a label with sufficient support replaces the current one; updated labels can take part in later iterations.

**Multi-joint assignment refinement.** Multi-joint configurations can expose assignments that single-joint excitation leaves ambiguous. With the axes, reference geometry and appearance fixed, we pose and render the Gaussians and continue optimizing the continuous assignment scores against the training images; anchored parts keep their assignments, and discrete labels are taken after optimization.

**Footprint-overflow pruning.** Samples from each Gaussian's $3\sigma$ extent are projected into multiple training views, and Gaussians whose fraction outside the foreground silhouette exceeds the overflow threshold are removed.

### 3.3 Implementation for each robot

**Panda.** The moving-set method jointly estimates image-supervised axes and moving scores (Section 4.4), and serial-chain ancestry gives the initial link labels, with link6 constrained by its part mask. Multi-joint assignment refinement is followed by reprojection voting, with the image-supervised axes and link6 labels fixed; other Gaussians may be reassigned to non-adjacent links. Finally, the image-supervised axes are replaced by the metric axes identified in Section 4.1, and the version that retains the image-supervised axes forms the axis-substitution pair.

**Laikago.** With the metric axes fixed, we optimize continuous link weights, take discrete labels and apply neighbour-majority smoothing. Two correction rounds, using 128 and 384 training views, each combine reprojection voting and spatial neighbourhood correction. Voting uses seven samples per Gaussian and restricts candidates to the same leg or the trunk; trunk labels stay fixed, and unresolved fragments are assigned to the trunk. Multi-joint training images then refine the assignment scores with calf scores fixed. Gaussians in the hip classes are merged into the adjacent thigh classes, leaving nine visible parts (the trunk and two segments per leg), and footprint-overflow pruning completes the representation.

**JAKA.** With the six metric axes fixed (Section 4.1), single-joint images fit the moving sets. All modelled Gaussians are downstream of the first joint, so moving scores are fitted for the remaining joints, and serial ancestry gives discrete labels. We then order the Gaussians by shortest-path distance on a reference-configuration neighbourhood graph and locate link regions from sustained label proportions; upstream labels inside distal regions are corrected, and residual upstream labels in the terminal region are assigned to the end link. The final model has six visible link classes. Graph construction, region detection and thresholds are given in Supplementary Note 2.

Multi-joint training images then adapt the Gaussian RGB colours, represented by the spherical-harmonic DC component, while centres, scales, rotations, opacities, link labels, axes and cameras stay fixed. The loss combines full-image mean-squared error, colour error inside the overlap of predicted and observed silhouettes (obtained by 7 × 7 erosion) and a quadratic penalty relative to the initial colours, with weights 0.35, 1 and 0.02; Adam runs for 1,200 updates with a learning rate of 0.008.

## 4 · Joint screw-axis estimation and metric identification

Revolute metric axes are identified from joint readings and end-effector observations (Section 4.1). For Laikago and JAKA, these axes are fixed before the Gaussian assignments are learned, whereas Panda first estimates axes and assignments jointly from images (Section 4.4) and then substitutes the metric axes. A separate set of gripper-opening images gives the prismatic direction (Section 4.2).

### 4.1 Identifying metric axes from relative end-effector motion

**Observations.** For each joint, we hold the other joints fixed and vary only $q_j$, recording 6–25 end-effector poses. The observed pose has the form

$$T(q_j) = A_j \exp(\hat{\eta}_j q_j) B_j. \qquad (2)$$

$A_j$ is the fixed transform upstream of joint $j$, and $B_j$ collects fixed downstream transforms to the observed end-effector frame, including marker mounting. The relative motion between configurations $a$ and $b$ is

$$\Delta T_{ab} = T(q_j^b) T(q_j^a)^{-1} = A_j \exp[\hat{\eta}_j (q_j^b - q_j^a)] A_j^{-1}. \qquad (3)$$

The downstream transform $B_j$ cancels. The axis in the observation frame satisfies $\hat{\xi}_j = A_j \hat{\eta}_j A_j^{-1}$. With the other joints at the reference configuration, this is the axis used in equation (1).

**Closed-form solution.** Let $(R_a, \mathbf{t}_a)$ be the reference pose and $(R_b, \mathbf{t}_b)$ another observed pose. Relative rotation and translation are

$$R^{\mathrm{obs}} = R_b R_a^{\top}, \qquad \mathbf{t}^{\mathrm{obs}} = \mathbf{t}_b - R^{\mathrm{obs}} \mathbf{t}_a.$$

The axis direction is unchanged by the relative rotation. We stack $(R^{\mathrm{obs}} - I)\boldsymbol{\omega} = \mathbf{0}$ over reference–observation pairs and estimate the unit direction by singular value decomposition. Its sign is chosen so that the observed rotation agrees with the encoder angle change. We reconstruct $R^{\mathrm{enc}} = \exp([\boldsymbol{\omega}]_{\times} \Delta q_j)$ and stack

$$(I - R^{\mathrm{enc}})\mathbf{p} = \mathbf{t}^{\mathrm{obs}}$$

for a least-squares estimate of an axis point. Points differing along the axis describe the same line. We use $\mathbf{p}_{\star} = \mathbf{p} - (\boldsymbol{\omega}^{\top}\mathbf{p})\boldsymbol{\omega}$, the point closest to the origin, as its representative. The procedure uses relative-motion forms of PoE calibration and screw-axis identification [38,39–42]

### 4.2 Learning the gripper translation direction from opening images

**Parameters and motion.** Panda's two fingers translate in opposite directions along a shared unit vector $\mathbf{a}$. Let $d_k$ be the recorded per-finger displacement. Gaussian $i$ has a sign $s_i \in \{-1, 0, +1\}$ determined by its left-finger, right-finger or non-finger assignment. With the arm at the reference configuration,

$$\boldsymbol{\mu}_i(d_k; \mathbf{a}) = \boldsymbol{\mu}_i^0 + s_i d_k \mathbf{a}, \qquad \| \mathbf{a} \|_2 = 1.$$

Translation preserves covariance, colour and opacity. Reference Gaussians, finger assignments and cameras are fixed, and the shared direction is fitted.

**Paired image loss.** Five openings are captured at each viewpoint. Each nonzero-opening image is paired with the closed-gripper image from the same viewpoint. Observed and predicted differences are

$$\Delta I_k = I_k(d_k) - I_k(0), \qquad \Delta \hat{I}_k(\mathbf{a}) = \hat{I}_k(d_k; \mathbf{a}) - \hat{I}_k(0).$$

The moving region is obtained by thresholding the RGB-channel mean absolute difference between the maximum-opening and closed images at $5/255$. Dilating that region by 12 pixels gives a local mask $M_k$. Expanding the original moving-region bounding box by 20 pixels gives a crop $B_k$. Writing the regional mean absolute value as $\langle|\cdot|\rangle$, define

$$E_{\Delta,k}(R; \mathbf{a}) = \langle |\Delta \hat{I}_k(\mathbf{a}) - \Delta I_k| \rangle_R,$$

$$E_{\mathrm{open},k}(R; \mathbf{a}) = \langle |\hat{I}_k(d_k; \mathbf{a}) - I_k(d_k)| \rangle_R.$$

The loss for a pair is

$$\ell_k(\mathbf{a}) = E_{\Delta,k}(M_k;\mathbf{a}) + 0.5E_{\Delta,k}(B_k;\mathbf{a}) + 0.05E_{\text{open},k}(B_k;\mathbf{a}).$$

Views with fewer than five changed pixels are excluded from direction fitting. Of 40 acquired viewpoints, 34 remain. A fixed random split assigns 26 to training and eight to validation, producing 104 and 32 nonzero-opening pairs.

**Direction search.** Initial directions comprise the 26 normalized nonzero vectors with components in $\{-1,0,1\}$, 96 Fibonacci-sphere directions, and both signs of the line between finger-Gaussian centroids. Duplicate directions are removed. Maximum-opening pairs from 24 training views screen candidates by cropped difference error. The best five directions each undergo 400 Adam updates on the training pairs, with learning rate 0.005 and gradient-norm limit 1, followed by renormalization after each update. Validation crop-difference error selects the final direction, with local-difference error breaking ties. Candidate evaluation weights views by their original number of changed pixels.

**Composition with the arm.** For a general arm configuration, finger translation is applied in the reference configuration and then carried by the end-link transform. Evaluation uses independently generated arm configurations and continuously sampled openings (Section 6.1).

### 4.3 Observation requirements for simultaneous calibration

We conduct a separate Panda experiment to examine simultaneous estimation of all seven axes and two fixed transforms (Supplementary Table 4). Each observation pairs a joint configuration with an end-effector pose. The transforms $G$ and $H$ relate the model to the measurement frame and describe marker mounting. Joint optimization fits the predicted end-effector poses to these observations.

Joint zero offsets are fixed at zero and the gripper remains stationary. Each revolute axis has four effective degrees of freedom, two for direction and two for line position. $G$ and $H$ each have six. A common change of model frame introduces six redundant coordinate degrees of freedom, giving

$$7 \times 4 + 6 + 6 - 6 = 34$$

effective unknowns. One end-effector pose provides at most six independent constraints. Parameter counting therefore requires $6M \geq 34$, or at least six poses. The sufficiency of configuration excitation is tested experimentally. Observation-count tests use nested subsets, with ten independent resamplings at selected counts. Position, orientation and combined-noise tests use four levels, repeated ten times at four observation counts. A parameter-group experiment progressively releases fixed transforms, axis points, directions and zero offsets. These analyses use simulator end-effector observations. The final self-model axes are obtained by per-joint identification.

### 4.4 Image-supervised axis estimation

**Joint parameterization.** For each joint, we estimate its unit direction, a point on the axis and the continuous Gaussian moving scores. Excitation image $k$ uses $\exp(\hat{\xi}_j \Delta q_{j,k})$, with the encoder angle relative to the reference. All images of that joint share one axis.

**Optimization.** Each of six initial directions is optimized for 200 steps, and training-image loss selects a candidate. Its direction and axis point initialize 1,500 further joint-optimization steps, with the moving scores reinitialized. Gradients of the image reconstruction loss pass through differentiable rendering to the axis and scores.

**Distal constraints and axis-point refinement.** For the last two Panda joints, we use the upstream moving set to restrict the image-loss region. A per-view bounding box encloses its rendered silhouette, with a margin covering wrist and hand motion. Downstream moving scores are multiplied by the upstream hard mask, so the downstream set is a subset of the upstream set. The axis point is initialized from the upstream estimate. After joint optimization, the direction is fixed and scores are thresholded at 0.5. With that mask fixed, only the axis point is optimized for 400 steps at learning rate 0.02. The resulting image-supervised axes are used in Panda model construction and the axis-substitution pair defined in Section 6.2.

## 5 · Jacobian and model-based control

### 5.1 Analytic Jacobian

Differentiating equation (1) gives the geometric Jacobian, with linear velocity in the first three rows and angular velocity in the last three. For revolute joint $j$, let $\boldsymbol{\omega}_j(\mathbf{q})$ and $\mathbf{c}_j(\mathbf{q})$ be the direction of its current axis and a point on it, obtained through the upstream transforms. For the end-effector position $\mathbf{p}_{EE}(\mathbf{q})$,

$$J_{1:3,j} = \boldsymbol{\omega}_j(\mathbf{q}) \times \left(\mathbf{p}_{EE}(\mathbf{q}) - \mathbf{c}_j(\mathbf{q})\right), \qquad J_{4:6,j} = \boldsymbol{\omega}_j(\mathbf{q}). \qquad (4)$$

For a prismatic joint, $J_{1:3,j} = \mathbf{v}_j(\mathbf{q})$ and $J_{4:6,j} = 0$, and columns outside the end-effector's ancestor path are zero. The analytic Jacobian is checked against finite differences (Supplementary Table 4).

### 5.2 Inverse kinematics

Given a task error $\mathbf{e}$ and task Jacobian $J_{\text{task}}$, damped least squares computes [43,44]

$$\Delta\mathbf{q} = J_{\text{task}}^{\top}(J_{\text{task}}J_{\text{task}}^{\top} + \lambda^2 I)^{-1}\mathbf{e}. \qquad (5)$$

Position reaching uses $\mathbf{e} = \mathbf{p}^{\star} - \hat{\mathbf{p}}(\mathbf{q})$ with the position Jacobian, and manipulation adds the gripper approach-direction residual and its Jacobian. Panda uses $\lambda = 10^{-3}$, a joint-update norm limit of 0.5 rad, at most 300 iterations and a position-residual threshold of $10^{-6}$ m. Each target is solved from eight initial configurations, including the joint-range midpoint, zero and random configurations, and the lowest-residual solution is retained. The Adam comparison uses eight random starts with 400 updates each. JAKA initialization, bounds and feedback are given in Section 5.5.

### 5.3 Panda reaching and manipulation

**World-coordinate reaching.** With random seed 0, we sample 200 targets uniformly over $x \in [0.35, 0.60]$ m, $y \in [-0.25, 0.25]$ m and $z \in [0.25, 0.60]$ m. Given the reference end-effector point $\mathbf{p}_0$, the predicted end-effector position is

$$\begin{bmatrix} \hat{\mathbf{p}}(\mathbf{q}) \\ 1 \end{bmatrix} = T_{l_{EE}}(\mathbf{q}) \begin{bmatrix} \mathbf{p}_0 \\ 1 \end{bmatrix}. \qquad (6)$$

For each target, inverse kinematics produces a joint command, which is executed in physical simulation before the end-effector position is read. The metric-anchored and image-supervised models, and the damped-least-squares and Adam solvers, share the same targets, reference point and evaluation conditions.

**Opening selection.** The two fingers carry separate Gaussian labels, and the finger-pad gap is the current distance between their centroids minus that distance in the fully closed state. Given the object diameter, the learned translation direction determines the opening at which this gap matches the diameter; the fixed opening is 16 mm per finger. The model-space evaluation has twelve sets, each combining three object positions and three ball radii sampled over 5–35 mm, giving 108 conditions; the criteria are defined in Section 6.3.

**Picking, carrying and placing.** Each task proceeds through approach, descent, closure, lift, transport, descent and release, and every phase uses the same kinematic model and checks gripper clearance above the table. At release, the ball bottom is 4 mm above the table. Separate calibration trials (seed 0) estimate the mean horizontal displacement caused by falling and rolling, which is subtracted from the release targets in evaluation (seed 1). Both axis models use the same 50 starts, goals, initial configurations and solver settings, with a horizontal release compensation of (−2.8, +17.3) mm; the same tasks are also evaluated with the compensation set to zero.

### 5.4 Laikago foot reaching and walking

**Foot reaching.** Each leg uses its own ancestor chain, with the foot point defined as the toe-node origin. Fifty reachable foot positions are sampled without replacement per leg, giving 200 targets. With the body fixed and gravity disabled, the inverse-kinematics solutions are set directly in the simulator before the foot positions are read.

**Standing and gait.** The four foot targets are translated together in the horizontal plane to centre them below the body centre of mass, a longitudinal shift of −124.9 mm, and target heights account for the 0.03 m radius of the spherical feet. A static crawl gait lifts one leg at a time, shifting the body into the support region of the other three legs before each lift. Swing feet follow cycloidal trajectories with zero endpoint velocity, in the order front left, rear right, front right and rear left. Standing height is 0.30 m, step length 0.06 m, foot clearance 0.05 m and lateral body sway ±0.04 m, and each of four cycles contains four phases of 25 control steps.

**Physical execution.** The body moves freely under gravity and contact. After the robot settles into a stable stance, the planned foot trajectories are translated to the settled start; at each control step, the desired body pose determines the four foot targets, and the branch models solve the joint commands. A proportional joint-position controller uses gain 0.1, with default friction 1.0 and a torque limit of 500 N·m. The simulator step is 1/240 s, with 60 substeps per control update, and a fixed offset of (0, −0.0438, 0.030) m relates the model base to the simulator centre-of-mass coordinates. Axis-substitution comparisons hold the gait, solver and dynamics settings fixed. The sensitivity analysis combines friction coefficients of 0.6, 0.8, 1.0 and 1.2 with torque limits of 40, 100, 200 and 500 N·m, giving 16 conditions.

### 5.5 JAKA reaching with marker feedback

**Registration.** With the joint axes fixed, end-effector marker poses at 20 calibration configurations determine the registration. For model poses $(R_k^{\text{mod}}, \mathbf{p}_k^{\text{mod}})$ and observations $(R_k^{\text{obs}}, \mathbf{p}_k^{\text{obs}})$, $R_H$ is the rotation mean of $R_k^{\text{obs}}(R_k^{\text{mod}})^{\top}$, and

$$\mathbf{t}_H = \frac{1}{M}\sum_{k=1}^{M}(\mathbf{p}_k^{\text{obs}} - R_H\mathbf{p}_k^{\text{mod}}).$$

Registered positions are $\hat{\mathbf{p}}(\mathbf{q}) = R_H\mathbf{p}^{\text{mod}}(\mathbf{q}) + \mathbf{t}_H$, and the position Jacobian is left-multiplied by $R_H$. Calibration configurations are separate from evaluation targets, and each fixed registration is applied to all predictions and control computations in its observation group.

The nominal baseline uses the same calibration observations to estimate a frame registration and a fixed tool-to-marker offset $\mathbf{d}$. Given the manufacturer's tool pose $\left(R_T(\mathbf{q}), \mathbf{p}_T(\mathbf{q})\right)$, the marker position is

$$\hat{\mathbf{p}}_{\text{nom}}(\mathbf{q}) = R_B(\mathbf{p}_T(\mathbf{q}) + R_T(\mathbf{q})\mathbf{d}) + \mathbf{t}_B,$$

where twelve rounds alternating rigid point-set registration with linear least squares for $\mathbf{d}$ give $R_B$, $\mathbf{t}_B$ and $\mathbf{d}$.

**Targets and initial configurations.** Fifteen targets are selected by marker visibility and measured displacement, five from each of three predefined displacement intervals. The robot moves to each generating configuration to measure the target position $\mathbf{p}^\star$; this configuration is retained for forward-kinematic evaluation but is not added to the inverse-kinematics start library. Starts comprise the 20 calibration configurations, zero and the centre configuration $(0,0,-5,0,0,0)°$; for each model, they are ranked by predicted distance to the target, and the ten nearest distinct configurations are used.

**Inverse kinematics and open-loop execution.** Joint coordinates are bounded to ±20° around zero, and the objective is the Euclidean distance between the predicted marker position and the target. The self-model uses its analytic position Jacobian with adaptive damped least squares (initial damping $7 \times 10^{-4}$, at most 600 iterations per start, joint updates limited to 1°), and backtracking accepts residual-reducing updates. The nominal baseline uses a forward-difference position Jacobian (increment $2 \times 10^{-4}$ rad, initial damping $10^{-3}$, at most 180 iterations per start). The lowest-residual solution is retained and executed only if its residual is at most 1 mm. The robot moves from the centre configuration along a joint-space path whose consecutive waypoints differ by at most 8° per joint, and after a 0.7 s settling period the joint readings and external marker position are recorded. Model execution order rotates across targets.

**Marker feedback.** After a valid open-loop execution of the self-model, the measured error $\mathbf{e}_k = \mathbf{p}^\star - \mathbf{p}_k^{\text{obs}}$ drives

$$\Delta\mathbf{q}_k = g_k J_k^\top (J_k J_k^\top + \lambda_c^2 I)^{-1}\mathbf{e}_k,$$

where $J_k$ is the current model position Jacobian, $\lambda_c = 1.2 \times 10^{-3}$ and the initial gain is $g_0 = 0.55$. Per-joint increments are limited to 0.75°, 0.40° and 0.20° for errors above 10 mm, between 3 and 10 mm, and at most 3 mm, respectively. An update is accepted when the measured error decreases by at least 0.5 mm, after which the gain is multiplied by 1.05 up to 0.65; otherwise, the robot returns to the last accepted configuration and the gain is halved. Missing marker measurements likewise trigger a return and a gain reduction. The procedure stops after ten attempts, eight accepted updates, an error of at most 0.8 mm, or a gain below 0.1, and the final error is the median of three independent measurement windows after stopping. The nominal baseline is evaluated only for forward prediction and open-loop reaching.

## 6 · Evaluation protocols

### 6.1 Image quality and configuration generalization

**Image metrics.** RGB values are normalized to [0,1], and per-frame PSNR is

$$\text{PSNR} = -10\log_{10}\left(\frac{1}{3HW}\sum_{u,v,c}\left(\hat{I}_{uvc} - I_{uvc}\right)^2\right).$$

We also compute the mean absolute pixel error, and threshold the predicted accumulated opacity and the observed foreground masks at 0.5 to obtain silhouette IoU and counts of extra and missing foreground pixels. Metrics are computed per frame and averaged arithmetically, with medians and quantiles reported separately. Panda and Laikago use their respective white-background test sets; the Panda test set, from reference [7], has 4,000 images of 2,000 distinct joint configurations, each recorded by two cameras.

**JAKA rendering.** The 31 test images come from one station at 800 × 800 resolution with a white background, and each prediction is generated directly from the recorded joint readings, cameras and fixed model parameters. PSNR uses floating-point RGB, SSIM uses an 11 × 11 uniform window averaged over the RGB channels, and LPIPS uses AlexNet with inputs in [−1, 1] [45]; silhouette centroid error is the pixel distance between the predicted and observed foreground centroids. Numerical fitting uses the training split, validation images determine the rendering parameters, and final test metrics are computed after all parameters are frozen (Supplementary Note 1).

**Ranges and combinations.** Joint-angle coverage is recorded separately for single-joint images, multi-joint training images and metric end-effector observations, and Panda's dedicated out-of-range set is assessed joint by joint against each source; the displayed configurations are selected by maximum joint angle. For JAKA, held-out configurations lie within the training-angle coverage, and single-joint and multi-joint groups are reported separately.

**Gripper composition.** A separate set of 4,000 configurations varies all seven arm joints and samples per-finger displacement continuously over 0–0.04 m. With the Gaussians, arm configuration and cameras fixed, we compare rendering at the recorded opening with rendering at zero opening within a shared region, the projected end-link

Gaussian bounding box at the recorded opening expanded by 16 pixels, and summarize the regional differences over five opening groups of 800 frames.

### 6.2 Joint axes and end-effector prediction

**Axis geometry.** Direction error is the smaller angle between the estimated and reference axis lines. For Panda and Laikago, line-position error is the shortest distance between the estimated and simulator-reference axes, averaged over seven and twelve joints, respectively.

**Arm and foot positions.** Panda's reference end-effector point is the centroid of the end-link Gaussians. We sample 200 configurations uniformly within the seven joint limits with seed 0, independently of axis fitting, and compare the positions predicted by equation (6) with the simulator. For Laikago, multi-joint records are sampled without replacement and each toe-node origin is evaluated.

**JAKA prediction.** Joint readings and external marker positions are recorded at 15 evaluation configurations, and the self-model and nominal baseline predict the same marker point from the same readings. For configuration $k$,

$$e_{\mathrm{FK},k} = \|\hat{\mathbf{p}}(\mathbf{q}_k) - \mathbf{p}_k^{\mathrm{obs}}\|_2,$$

where $\hat{\mathbf{p}}$ includes the registration of Section 5.5. Mean, median and 90th-percentile errors are reported over all 15 configurations.

**Controlled axis substitution.** The Panda axis-substitution pair consists of the image-supervised and metric-anchored models, which share Gaussians, assignments, cameras, reference end-effector point and evaluation configurations and differ only in their axes. Both models are rendered against white and black backgrounds and predict end-effector positions at the same configurations. For Laikago's single-joint loss comparison, the reference Gaussians and moving sets are fixed and either the simulator axes or the image-supervised axes are substituted, with each joint using the same ten excitation images and loss function.

### 6.3 Control tasks

**Reaching error.** Reaching error is the Euclidean distance from the executed end-effector position to the target: for Panda, the position after execution in physical simulation; for Laikago, the position after setting the solved configuration; and for JAKA, an external marker measurement. The in-model residual is computed separately from the model-predicted position.

**Gripper geometry.** A condition passes when the finger-midpoint error is at most 2 mm, the approach-direction error from vertical is at most 2°, the predicted gap is within 1 mm of the object diameter and the opening is within the physical travel. Adaptive and fixed opening use identical object sizes and positions.

**Manipulation success.** Both fingers must contact the object, lift it by at least 20 mm and keep its displacement relative to the gripper within 25 mm during transport; after release, it must remain on the table with a height error of at most 15 mm and a planar error of at most 40 mm. The arm must not contact the table, and the maximum joint-tracking error must not exceed 0.03 rad. Grasp-point error is the distance from the finger midpoint to the ball centre at closure, and placement error is measured in the table plane after the object settles.

**Walking.** We record forward displacement, lateral drift, maximum tilt and falls for the metric-anchored and image-supervised models across the 16 friction and torque settings.

**JAKA inverse kinematics and open-loop reaching.** All 15 targets form the denominator for inverse-kinematics attempts, and accepted solutions are executed without task-space feedback, giving

$$e_{\mathrm{open},k} = \|\mathbf{p}_{k,\mathrm{open}}^{\mathrm{obs}} - \mathbf{p}_k^{\star}\|_2.$$

Open-loop statistics include targets with executable solutions and valid measurements, with sample counts stated; paired model comparisons use targets completed by both models, and target-level records retain inverse-kinematics failures.

**JAKA feedback reaching.** The initial error $e_{\mathrm{init},k}$ is measured again before feedback, and the final error $e_{\mathrm{final},k}$ is the median of three measurement windows after stopping (Section 5.5). We compare the initial and final distributions, count targets whose error decreases and count final errors within 5 mm and 10 mm, reporting the corresponding execution denominators (Supplementary Table 7).

# Figures

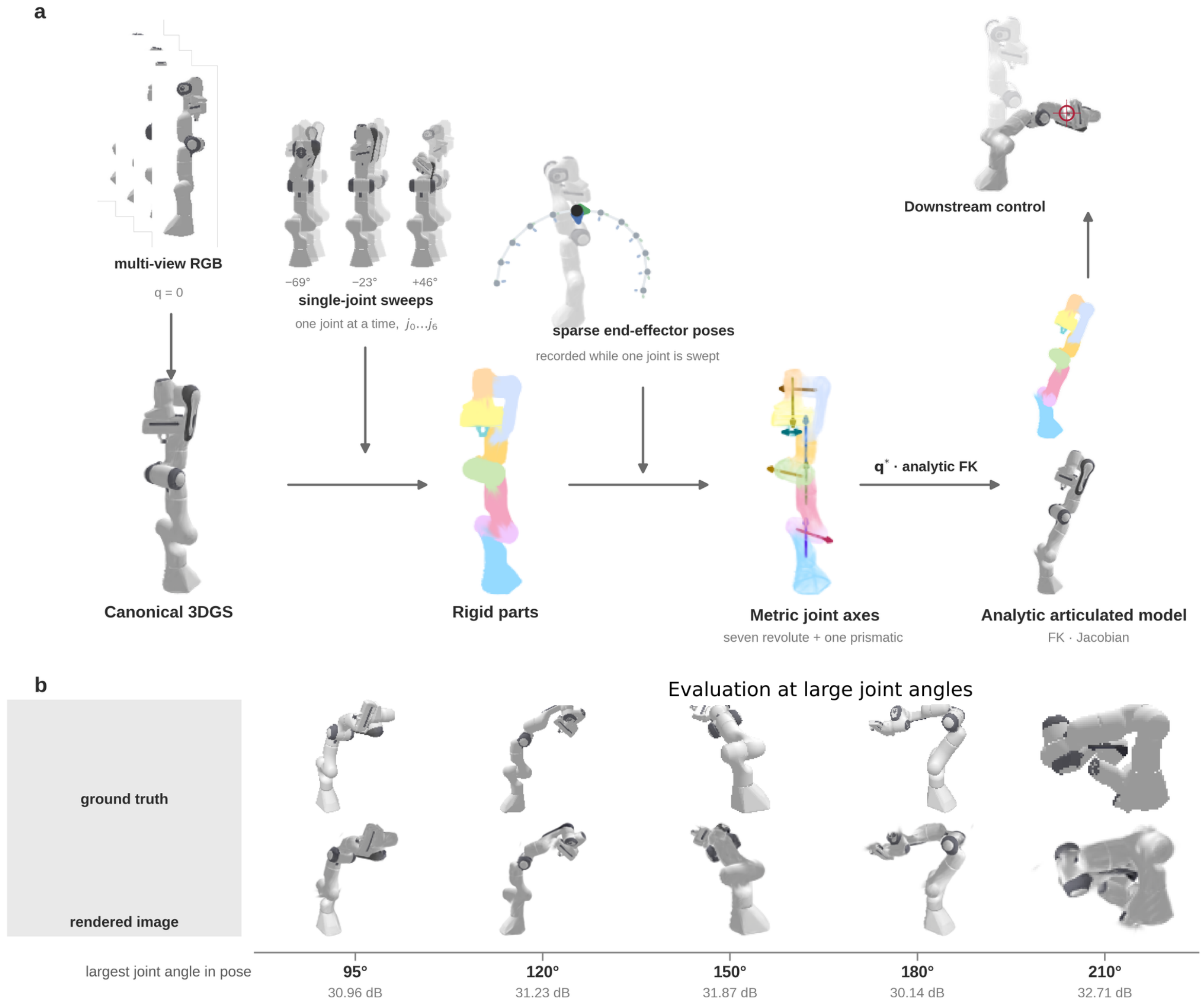

**Fig. 1 | Analytic self-modelling and evaluation at large joint excursions.** a, Multi-view RGB images at the reference configuration reconstruct static Gaussians. Single-joint images provide link assignments, sparse end-effector poses identify metric revolute axes, and separate gripper images identify a prismatic direction. The model supplies forward kinematics and an analytic Jacobian. b, Five representative Panda configurations from the out-of-range test. Single-joint images cover approximately ±68.8°, and multi-joint training angles reach approximately ±90°. End-effector observation ranges are specified per joint. The largest joint angle in the displayed configurations increases from 95° to 210°. Extrapolation is assessed separately for each joint and each fitting source (Supplementary Table 1b). The upper row shows reference images and the lower row model renderings.

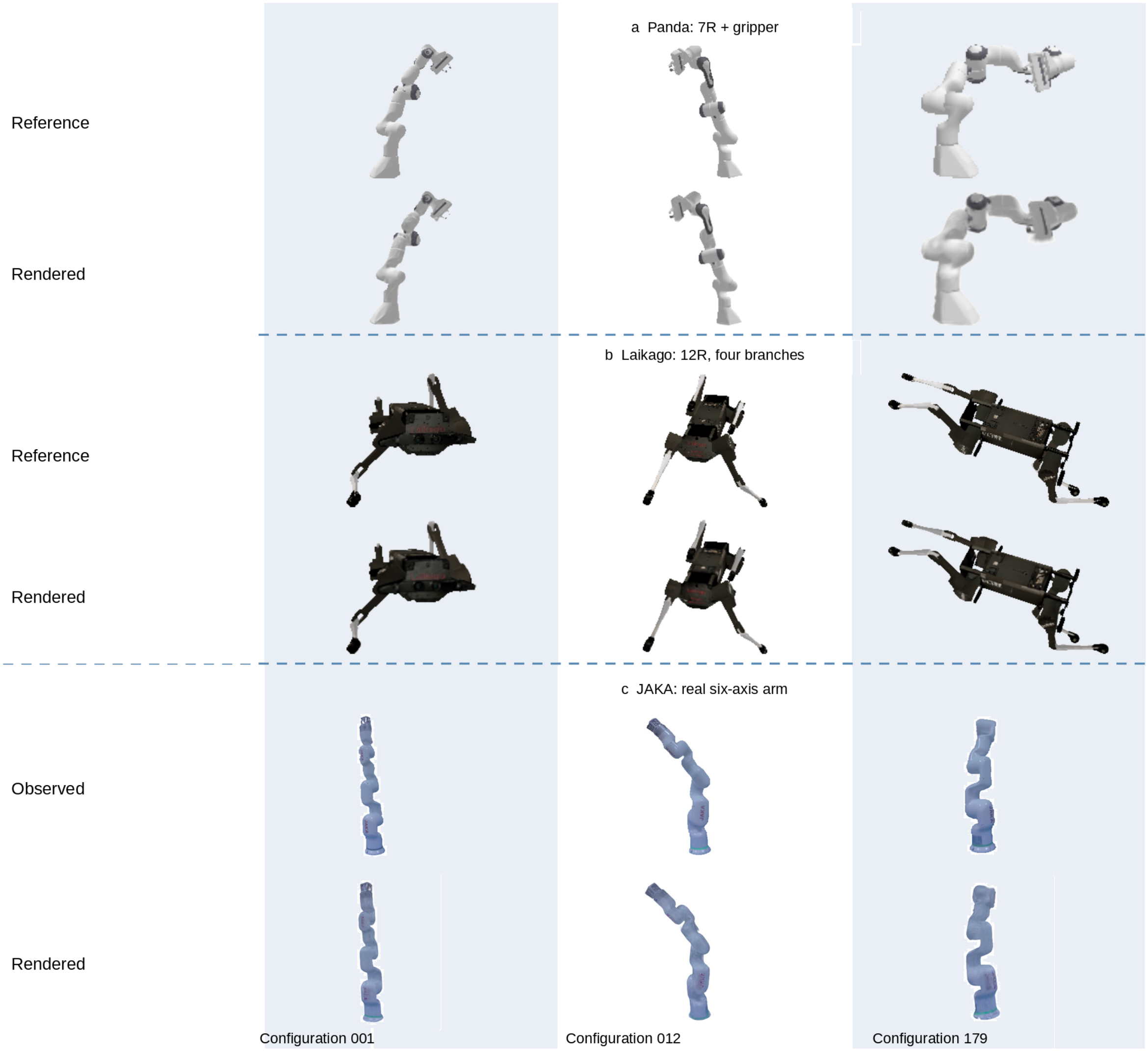

**Fig.** 2 | Observed and rendered appearance of three robot self-models. a, Panda with seven revolute joints and a gripper prismatic degree of freedom, reaching 31.8 dB mean PSNR over 4,000 test images. b, Laikago with twelve revolute joints on four branches, reaching 26.9 dB over 213 test images. c, The real six-axis JAKA driven by recorded encoder readings, reaching 31.0 dB over 31 test images at one camera station. Configurations 001, 012 and 179 are shown. Observed and rendered images share a crop for each configuration. JAKA full-image metrics are reported in Supplementary Table 2a.

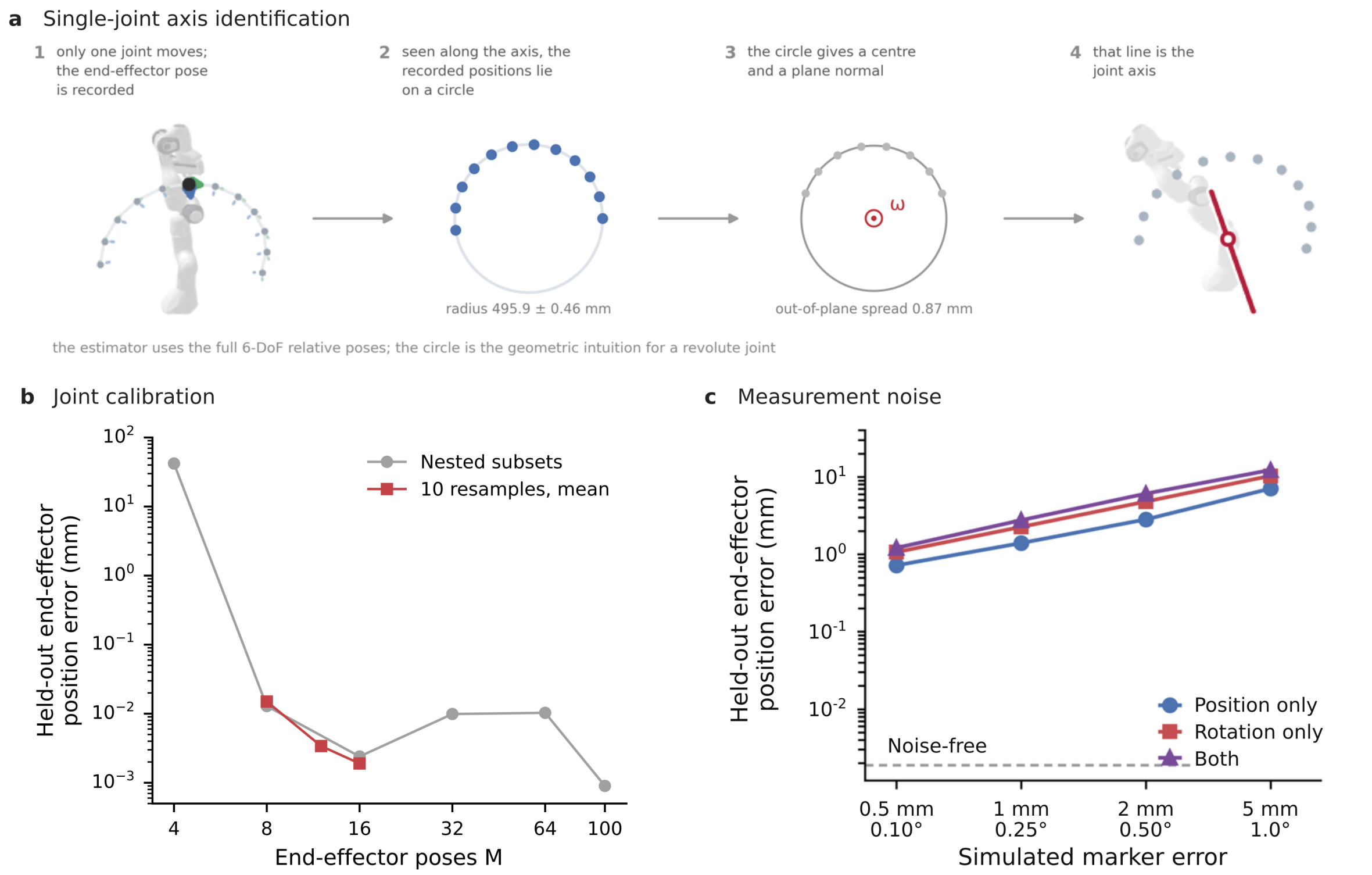

**Fig. 3 | Metric anchoring from sparse end-effector poses.** a, Under single-joint rotation, end-effector positions trace a circle whose centre lies on the axis and whose plane normal is parallel to it; the estimator itself uses the full relative rigid poses. b, Held-out end-effector error against observation count in joint calibration. Grey points use nested subsets, and red points show means over ten independent resamplings. c, Held-out error under simulated position, orientation and combined marker noise with $M = 16$. Each condition is repeated ten times. Combined noise levels of 1 mm in position and 0.25° in orientation give a held-out error of 2.75 mm. Panels b and c jointly estimate seven axes and two fixed transforms, separately from the per-joint identification used in the final model (Supplementary Table 4).

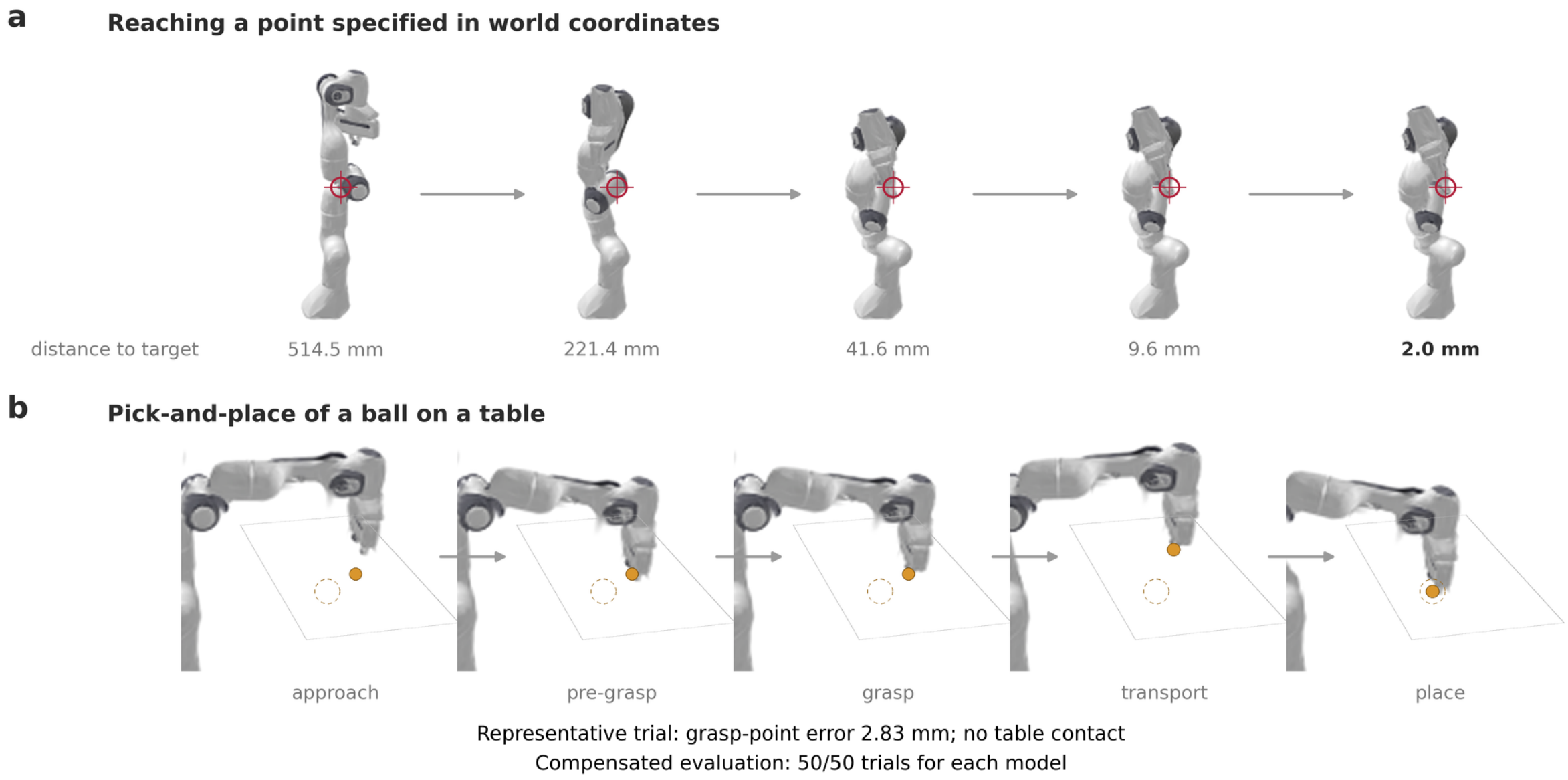

**Fig.** 4 | Manipulation with the analytic self-model. a, In a representative reaching trial, the distance to a world-coordinate target decreases from 514.5 mm to 2.0 mm. Across 200 targets, mean after-execution errors are 2.280 mm for the metric-anchored model and 7.636 mm for the image-supervised model. b, Picking, carrying and placing a ball. The displayed trial has a grasp-point error of 2.83 mm. With the same release-point compensation, both models complete all 50 trials. Mean grasp and placement errors are 2.78 and 3.21 mm for the metric-anchored model, compared with 13.18 and 19.16 mm for the image-supervised model (Supplementary Table 5).

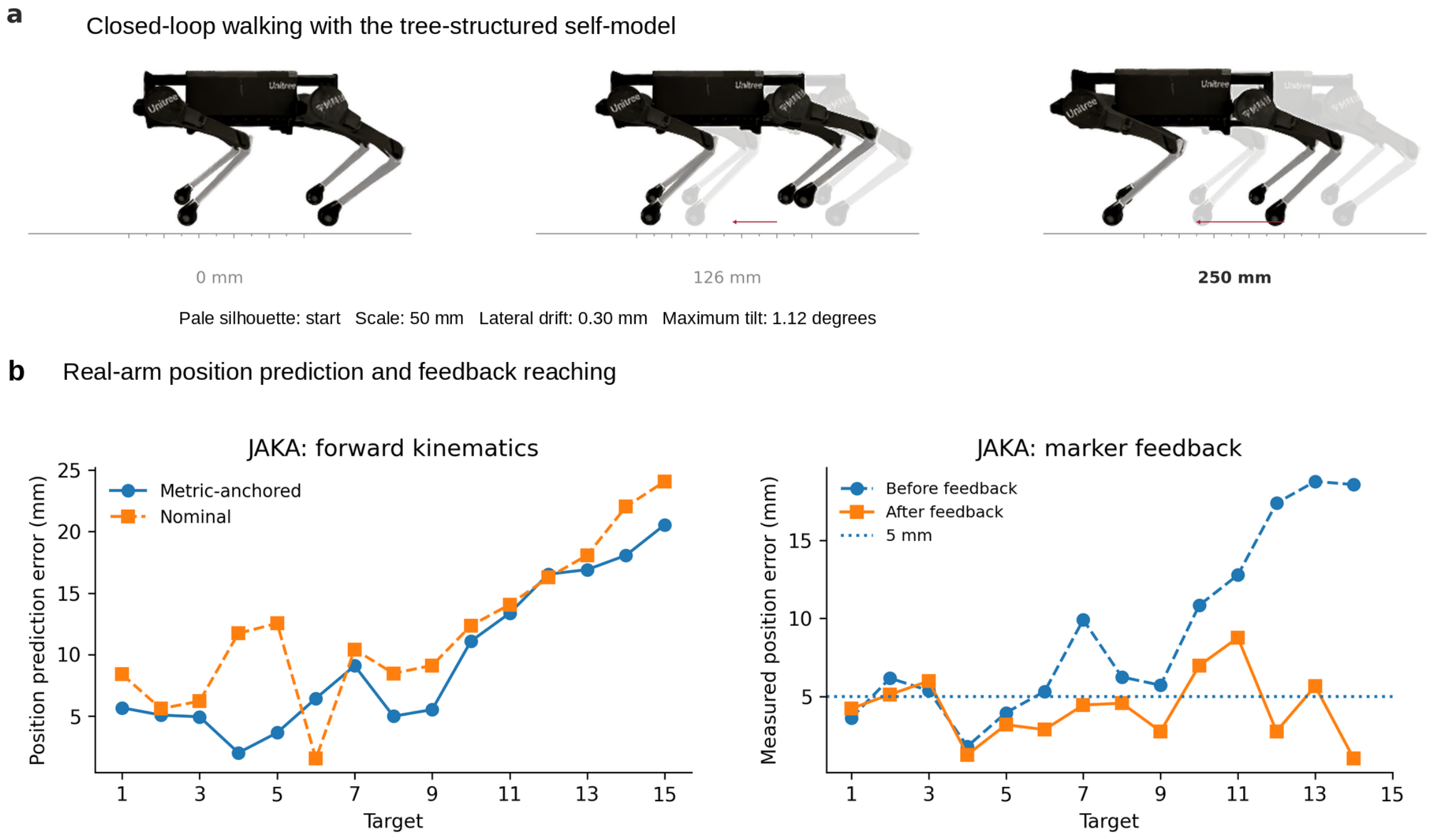

**Fig.** 5 | Tree-structured walking and real-arm feedback reaching. a, Laikago crawling in physical simulation with gravity and ground contact. The pale silhouette marks the starting configuration, and scale divisions are 50 mm. For the image-supervised and metric-anchored models, respectively, in-model residuals are 0.144 and 0.198 μm, median foot errors during walking are 79.15 and 8.35 mm, forward displacements are 151.9 and 250.1 mm, and maximum tilt angles are 11.80° and 1.12°. b, External measurements on JAKA. The left plot compares position predictions at all 15 targets. The right plot shows errors before and after feedback for the self-model's 14 valid executions. Target 15 has no executable inverse-kinematics solution and hence no feedback measurement. The dotted line marks 5 mm. Measurements are summarized in Supplementary Table 7.

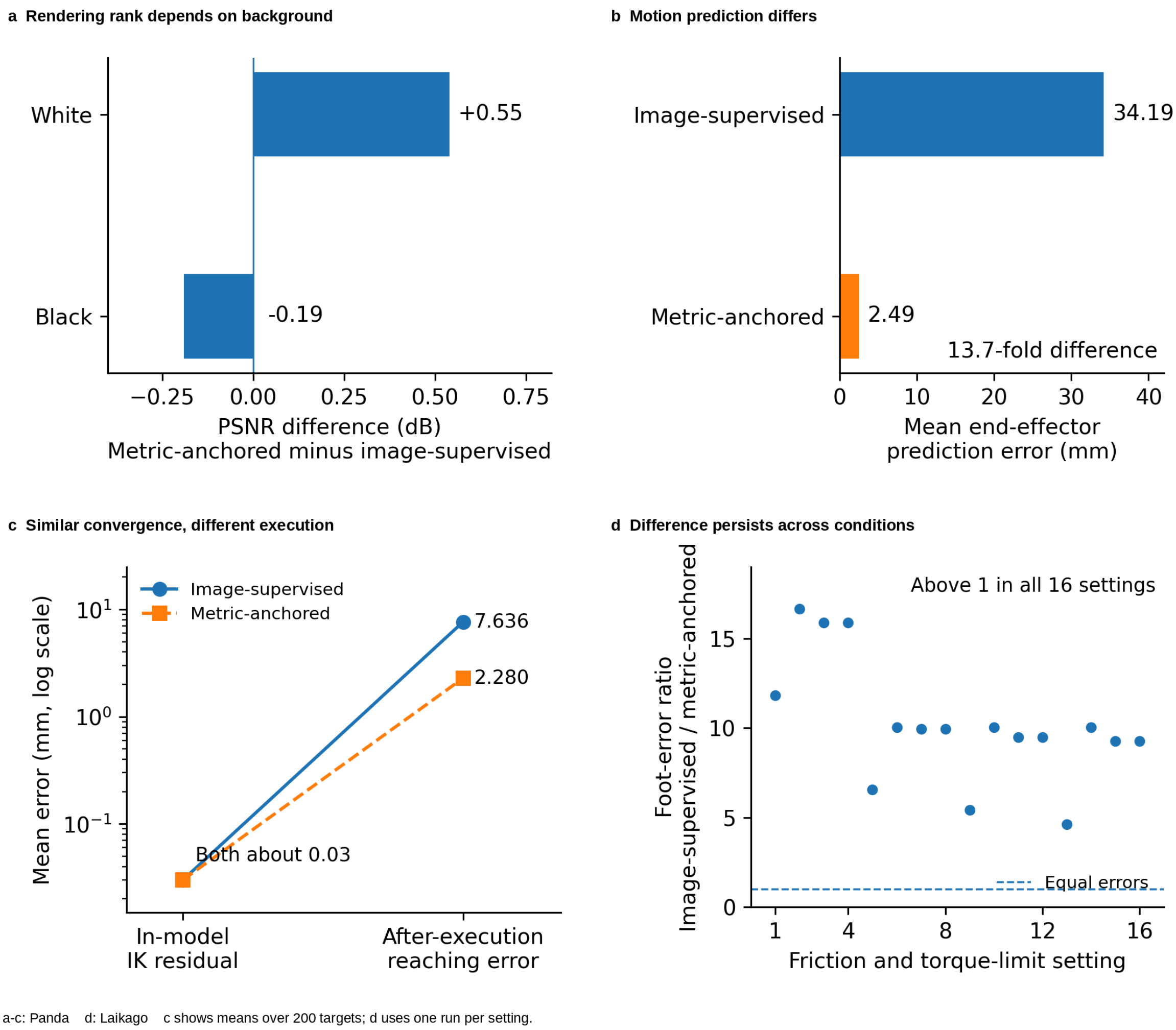

**Fig.** 6 | Internal fit measures and physical motion accuracy. a,b, Panda axis-substitution pair (Methods Section 6.2). a, Metric-anchored PSNR minus image-supervised PSNR is +0.55 dB against white and −0.19 dB against black. Background changes the rendering rank but not the kinematics. b, Mean end-effector prediction errors are 34.19 mm and 2.49 mm for the image-supervised and metric-anchored models, a 13.7-fold difference. c, In the separate 200-target reaching experiment, both models have a mean in-model residual of about 0.03 mm, yet mean execution errors are 7.636 mm and 2.280 mm. The vertical axis is logarithmic. d, The ratio of image-supervised to metric-anchored foot error during walking on Laikago is above one in all 16 settings. Settings are ordered by friction 0.6, 0.8, 1.0 and 1.2, each with torque limits 40, 100, 200 and 500 N·m. Each setting is evaluated once (Supplementary Tables 5a and 6b).

## References

[1] Bongard, J., Zykov, V. & Lipson, H. Resilient machines through continuous self-modeling. Science 314, 1118–1121 (2006).

[2] Kwiatkowski, R. & Lipson, H. Task-agnostic self-modeling machines. Sci. Robot. 4, eaau9354 (2019).

[3] Chen, B., Kwiatkowski, R., Vondrick, C. & Lipson, H. Full-body visual self-modeling of robot morphologies. Sci. Robot. 7, eabn1944 (2022).

[4] Díaz Ledezma, F. & Haddadin, S. Machine learning-driven self-discovery of the robot body morphology. Sci. Robot. 8, eadh0972 (2023).

[5] Hang, K., Bircher, W. G., Morgan, A. S. & Dollar, A. M. Manipulation for self-identification, and self-identification for better manipulation. Sci. Robot. 6, eabe1321 (2021).

[6] Hu, Y., Lin, J. & Lipson, H. Teaching robots to build simulations of themselves. Nat. Mach. Intell. 7, 484–494 (2025).

[7] Hu, K., Yu, P. & Tan, N. Learning high-fidelity robot self-model with articulated 3D Gaussian splatting. Int. J. Robot. Res. 45, 1478–1496 (2026). https://doi.org/10.1177/02783649251396980.

[8] Schulze, L. & Lipson, H. High-degrees-of-freedom dynamic neural fields for robot self-modeling and motion planning. In Proc. IEEE International Conference on Robotics and Automation 3064–3070 (IEEE, 2024).

[9] Li, S. L. et al. Controlling diverse robots by inferring Jacobian fields with deep networks. Nature 643, 89–95 (2025).

[10] Hu, Y., Chen, B. & Lipson, H. Egocentric visual self-modeling for autonomous robot dynamics prediction and adaptation. npj Robot. 3, 14 (2025).

[11] Liu, J., Mahdavi-Amiri, A. & Savva, M. PARIS: part-level reconstruction and motion analysis for articulated objects. In Proc. IEEE/CVF International Conference on Computer Vision 352–363 (IEEE, 2023).

[12] Liu, Y. et al. Building interactable replicas of complex articulated objects via Gaussian splatting. In Proc. International Conference on Learning Representations (2025).

[13] Guo, J. et al. ArticulatedGS: self-supervised digital twin modeling of articulated objects using 3D Gaussian splatting. In Proc. IEEE/CVF Conference on Computer Vision and Pattern Recognition 27144–27153 (IEEE, 2025).

[14] Lin, S. et al. SplArt: articulation estimation and part-level reconstruction with 3D Gaussian splatting. In Proc. IEEE/CVF International Conference on Computer Vision 8841–8851 (IEEE, 2025).

[15] Lewis, S., Chandra, V., Gao, T. & Jenkins, O. C. SPLATART: articulated Gaussian splatting with estimated object structure. Preprint at https://arxiv.org/abs/2506.12184 (2025).

[16] Kim, S., Ha, J., Kim, Y. H., Lee, Y. & Park, F. C. ScrewSplat: an end-to-end method for articulated object recognition. In Proc. 9th Conference on Robot Learning 309–335 (PMLR, 2025).

[17] Kim, S. & Jaquier, N. ChainSplat: a physics-inspired screw-theoretic model for learning deformable linear object dynamics from multi-view RGB videos. Preprint at https://arxiv.org/abs/2608.28570 (2026).

[18] Chen, Q. et al. URDFormer: a pipeline for constructing articulated simulation environments from real-world images. In Proc. Robotics: Science and Systems XX, 124 (2024).

[19] Mandi, Z., Weng, Y., Bauer, D. & Song, S. Real2Code: reconstruct articulated objects via code generation. In Proc. International Conference on Learning Representations (2025).

[20] Le, L. et al. Articulate-Anything: automatic modeling of articulated objects via a vision-language foundation model. In Proc. International Conference on Learning Representations (2025).

[21] Qiu, X. et al. Articulate AnyMesh: open-vocabulary 3D articulated objects modeling. In Proc. 9th Conference on Robot Learning 4585–4606 (PMLR, 2025).

[22] Liu, R., Canberk, A., Song, S. & Vondrick, C. Differentiable robot rendering. In Proc. 8th Conference on Robot Learning, Proceedings of Machine Learning Research 270, 117–129 (2025).

[23] Lou, H. et al. Robo-GS: a physics consistent spatial-temporal model for robotic arm with hybrid representation. In Proc. IEEE International Conference on Robotics and Automation 15379–15386 (IEEE, 2025).

[24] Lin, J. et al. AutoURDF: unsupervised robot modeling from point cloud frames using cluster registration. In Proc. IEEE/CVF Conference on Computer Vision and Pattern Recognition 27628–27637 (IEEE, 2025).

[25] Allegro, D., Terreran, M. & Ghidoni, S. Calib3R: hand-eye calibration and 3D metric-scaled scene reconstruction with 3D foundation models. Preprint at https://arxiv.org/abs/2509.08813v2 (2026).

[26] Yang, S. et al. Robo3R: enhancing robotic manipulation with accurate feed-forward 3D reconstruction. In Proc. Robotics: Science and Systems XXII, 56 (2026). https://doi.org/10.15607/RSS.2026.XXII.056.

[27] Mildenhall, B. et al. NeRF: representing scenes as neural radiance fields for view synthesis. Commun. ACM 65, 99–106 (2022). https://doi.org/10.1145/3503250.

[28] Zhang, K., Riegler, G., Snavely, N. & Koltun, V. NeRF++: analyzing and improving neural radiance fields. Preprint at https://arxiv.org/abs/2010.07492 (2020).

[30] Gao, J. et al. Relightable 3D Gaussians: realistic point cloud relighting with BRDF decomposition and ray tracing. In Proc. European Conference on Computer Vision 73–89 (Springer, 2024).

[31] Brockett, R. W. Robotic manipulators and the product of exponentials formula. In Mathematical Theory of Networks and Systems, Lecture Notes in Control and Information Sciences Vol. 58, 120–129 (Springer, 1984).

[32] Lynch, K. M. & Park, F. C. Modern Robotics: Mechanics, Planning, and Control (Cambridge Univ. Press, 2017).

[33] Carion, N. et al. SAM 3: segment anything with concepts. Preprint at https://arxiv.org/abs/2511.16719 (2025).

[34] Garrido-Jurado, S., Muñoz-Salinas, R., Madrid-Cuevas, F. J. & Marín-Jiménez, M. J. Automatic generation and detection of highly reliable fiducial markers under occlusion. Pattern Recognit. 47, 2280–2292 (2014).

[35] Coumans, E. & Bai, Y. PyBullet, a Python module for physics simulation for games, robotics and machine learning. http://pybullet.org (2016–2021).

[36] Wang, Z., Bovik, A. C., Sheikh, H. R. & Simoncelli, E. P. Image quality assessment: from error visibility to structural similarity. IEEE Trans. Image Process. 13, 600–612 (2004).

[37] Kavan, L., Collins, S., Žára, J. & O'Sullivan, C. Geometric skinning with approximate dual quaternion blending. ACM Trans. Graph. 27, 105 (2008).

[38] Chen, G., Wang, H. & Lin, Z. Determination of the identifiable parameters in robot calibration based on the POE formula. IEEE Trans. Robot. 30, 1066–1077 (2014).

[39] Okamura, K. & Park, F. C. Kinematic calibration using the product of exponentials formula. Robotica 14, 415–421 (1996).

[40] Chen, I.-M., Yang, G., Tan, C. T. & Yeo, S. H. Local POE model for robot kinematic calibration. Mech. Mach. Theory 36, 1215–1239 (2001).

[41] Li, C., Wu, Y., Löwe, H. & Li, Z. POE-based robot kinematic calibration using axis configuration space and the adjoint error model. IEEE Trans. Robot. 32, 1264–1279 (2016).

[42] Wang, H., Shen, S. & Lu, X. A screw axis identification method for serial robot calibration based on the POE model. Ind. Robot 39, 146–153 (2012).

[43] Nakamura, Y. & Hanafusa, H. Inverse kinematic solutions with singularity robustness for robot manipulator control. J. Dyn. Syst. Meas. Control 108, 163–171 (1986).

[44] Wampler, C. W. Manipulator inverse kinematic solutions based on vector formulations and damped least-squares methods. IEEE Trans. Syst. Man Cybern. 16, 93–101 (1986).

[45] Zhang, R., Isola, P., Efros, A. A., Shechtman, E. & Wang, O. The unreasonable effectiveness of deep features as a perceptual metric. In Proc. IEEE/CVF Conference on Computer Vision and Pattern Recognition 586–595 (IEEE, 2018).

# Supplementary Information

## Supplementary Note 1 · Data acquisition and preprocessing

Panda and Laikago RGB images are rendered in PyBullet at 400 × 400 pixels. Robot descriptions generate images and evaluation references but are not inputs to self-model fitting. Reference images establish static geometry, single-joint excitation provides motion evidence, and multi-joint images support refinement and evaluation. Data sizes and configuration ranges are summarized in Supplementary Table 1.

The 4,000 Panda test images correspond to 2,000 distinct joint configurations observed by two cameras; rendering metrics are averaged per image.

JAKA static reconstruction uses 497 training, ten validation and ten test images. Reference geometry is retained at step 20,000. Training uses a white background, spherical harmonics up to degree 3, and $0.8L_1 + 0.2(1 - \mathrm{SSIM})$ with an additional foreground-opacity $L_1$ term weighted by 0.1. Densification begins at step 500, runs every 100 steps and ends at step 15,000, with gradient threshold 0.0002 and opacity resets every 3,000 steps. The articulated representation uses RGB colours from the DC component.

JAKA RGB frames are captured at 1920 × 1080 and remapped to an 800 × 800 virtual pinhole camera. All multi-pose images come from one camera station. SAM 3 with connected-component cleanup provides the robot foreground, and the end-effector marker is excluded from the body foreground used for rendering evaluation. Split sizes and joint ranges are reported in Supplementary Table 1.

Numerical fitting uses the training split, and validation images select the rendering parameters. Test images were inspected during exploratory diagnostics but were not used for numerical fitting or final validation-based model selection; the reported test metrics were computed after all parameters were frozen.

## Supplementary Note 2 · Assignment and rendering settings

### Panda

Per-joint moving-set estimation jointly fits axis direction, axis point and moving scores. Each joint uses 30 training images for fitting and 20 held-out images for evaluation. Direction and point learning rates are 0.005, and moving-score learning rate is 0.05. The six-direction screening, joint fitting and axis-point refinement are described in Methods Section 4.4. The upstream-silhouette crop, moving-set subset constraint and upstream point initialization apply to the last two joints. The other five use the full image. A local part mask constrains the axisymmetric link6, followed by multi-joint assignment refinement and reprojection voting. Independently identified metric axes are then substituted.

### Laikago

Metric axes remain fixed while continuous link scores are optimized. Initial fitting uses 4,000 steps, learning rate 0.05 and score scale 0.35. A cosine schedule decreases temperature from 1.0 to 0.15. Image loss is $0.8L_1 + 0.2(1 - \mathrm{SSIM})$, with silhouette $L_1$ weighted by 1.0, both evaluated inside the foreground box. The eight-neighbour assignment regularizer is evaluated over all Gaussians with weight 0.02. After hardening, up to two smoothing passes relabel a Gaussian when at least six of its eight neighbours agree. Smoothing stops when no label changes.

Correction rounds use 128 and 384 training views. Each Gaussian contributes its centre and six $\pm 3\sigma$ principal-axis points. Candidates are from the same leg or trunk, and trunk labels remain fixed. The first silhouette threshold is 0.5 with two-pixel dilation. The second is 0.7 without dilation. Both rounds use seed 20260718.

| Criterion | Setting |
|---|---|
| Current-label score | At most 0.8 |
| Best candidate score | At least 0.9 |
| Improvement over current label | At least 0.15 |
| Gap to second-best label | At least 0.05 |
| Valid views | At least 80% and at least 16 |
| Relabelling cap per round | 3% of all Gaussians |
| Minimum resulting class | 200 Gaussians |
| Fragment size | Fewer than 80 Gaussians |
| Neighbourhood | Eight nearest neighbours |
| Fragment support | Four external neighbours sharing a label |
| Fragment-correction passes | Five |
| Unresolved fragments | Assigned to trunk |
| Multi-joint refinement | 3,000 steps, calf scores fixed |
| Footprint pruning | Overflow above 0.3 over 64 training views |

Gaussians assigned to hip classes are merged with the adjacent thigh classes after multi-joint refinement, leaving the trunk and two visible segments per leg. All twelve kinematic joints are retained. The correction ablation retains or skips the two voting and neighbourhood rounds. Both versions then undergo multi-joint refinement and are evaluated on the same 213 frames before hip merging and pruning (Supplementary Table 2b).

### Dual-quaternion blending

Candidate quaternion signs are aligned to the largest-weight candidate. Weighted sums of real and dual parts are divided by the norm of the blended real part. Rotation follows from the normalized real part. Translation is twice the vector part of the normalized dual part multiplied by the conjugate real part. This implements rigid-transform blending by dual quaternions [37]

### JAKA

Reference Gaussians are reconstructed under RGB and foreground-opacity supervision. With six metric axes fixed, excitation images estimate moving sets and serial ancestry gives discrete labels. All modelled Gaussians are downstream of the first joint. The final representation contains 190,731 Gaussians.

**Graph distance.** The supplied eight-neighbour graph connects Gaussian centres in the reference configuration. For vertex $i$, the local scale is the median distance to its eight neighbours, lower-bounded by $10^{-6}$ m. A directed neighbour entry is retained when it is finite, not a self-edge and no longer than three times this local scale. Each retained entry contributes its Euclidean distance in both directions. Coincident sparse-matrix entries are summed during assembly.

Principal-component coordinates of the Gaussian centres locate the base end. The first component is oriented from the median position of L1 towards that of L6. Base seeds are L1 Gaussians at or below the fifth percentile of their first-component coordinates. When more than 1,500 seeds qualify, 1,500 evenly spaced indices are retained. A virtual source connects to these seeds with weight $10^{-12}$. Undirected Dijkstra distances from this source define the ordering coordinate. Region correction requires at least 95% of the Gaussians to have finite distance. The principal component selects source seeds, whereas graph distance orders the link regions.

**Link regions.** Among finite-distance L3–L6 Gaussians, the 0.5th and 99.5th distance percentiles define 220 equal-width bins. Five-bin moving sums of label counts and total counts give smoothed label proportions. L4, L5 and L6 onsets are detected in order as the earliest three consecutive bins each with mass at least 40 and proportions at least 0.40, 0.40 and 0.30, respectively. L3 labels in the L4 region become L4. L3 and L4 labels in the L5 region become L5. Residual L3–L5 labels beyond the L6 onset become L6. L1 and L2 are unchanged.

Appearance and rendering parameters. Multi-pose colour adaptation updates only RGB/DC parameters. Full-image mean-squared error, overlap-interior colour error and the initial-colour quadratic penalty have weights 0.35, 1 and 0.02. A 7 × 7 erosion defines the overlap interior. Adam uses learning rate 0.008 for 1,200 updates. All other Gaussian, label, axis and camera parameters remain fixed.

## Supplementary Table 1 · Data and configuration ranges

### 1a · Model sizes and data splits

| Item | Panda | Laikago | JAKA |
|---|---|---|---|
| Platform and structure | Simulation, 7R + gripper, serial | Simulation, 12R, four branches | Hardware, 6R, serial |
| Reference Gaussians | 9,935 | 117,538 | 190,731 |
| Evaluation Gaussians | 9,935 | 117,476 | 190,731 |
| Articulated / multi-pose images, train / val / test | 10,000 / 1 / 4,000 | 1,587 / 200 / 213 | 146 / 25 / 31 |
| Rendering resolution | 400 × 400 | 400 × 400 | 800 × 800 |
| Metric observation target | End link | Toe node per leg | Rigid end-effector marker |

| Modelling observations | Panda | Laikago |
|---|---|---|
| Single-joint images, train / val / test | 210 / 1 / 140 | 424 / 61 / 55 |
| End-effector observations per joint | 6–12 angles | Hip / thigh / calf: 12 / 12 / 6 |
| Out-of-range rendering | 750 frames with reference images | Not evaluated |

### 1b · Panda joint ranges, in degrees

| Joint | Single-joint images | Multi-joint training | Metric observations | Out-of-range test |
|---|---|---|---|---|
| j0 | ±68.75 | [−89.99, 89.98] | [−150, 150] | [−162.54, 162.60] |
| j1 | ±68.75 | [−89.98, 89.98] | [−94.5, 94.5] | [−98.94, 98.89] |
| j2 | ±68.75 | [−89.91, 89.99] | [−150, 150] | [−162.57, 162.59] |
| j3 | ±68.75 | [−89.82, 0.82] | [−125, 0] | [−172.45, −4.16] |
| j4 | ±68.75 | [−89.95, 89.99] | [−150, 150] | [−162.48, 162.62] |
| j5 | ±68.75 | [−1.33, 89.95] | [−4.5, 150] | [−0.12, 210.68] |
| j6 | ±68.75 | [−89.96, 89.99] | [−150, 150] | [−162.60, 162.68] |

Panda joint indices are zero-based. Asymmetric limits produce asymmetric sampling for j3 and j5. Each joint has test values beyond every fitting-data interval in at least one direction.

### 1c · Laikago ranges, in degrees

| Data | Hip | Thigh | Calf |
|---|---|---|---|
| Reference | 0 | 0 | 0 |
| Single-joint images | [−44.52, 54.50] | [−61.02, 90.00] | [−90.00, 0.00] |
| Multi-joint images | [−40.01, 47.98] | [−59.57, 155.80] | [−99.19, −0.03] |
| Metric observations | [−45.02, 54.00] | [−67.04, 150.00] | [−111.59, 0] |

The front-right leg is shown. Corresponding ranges are similar across legs.

### 1d · JAKA multi-pose image ranges, in degrees

| Joint | Training | Validation | Test |
|---|---|---|---|
| J1 | [−180.00, 180.00] | [−153.00, 180.00] | [−153.50, 120.00] |
| J2 | [−50.00, 50.00] | [−31.48, 44.22] | [−31.23, 42.50] |
| J3 | [−43.71, 50.00] | [−50.00, 30.00] | [−42.50, 42.74] |
| J4 | [−180.00, 180.00] | [−180.00, 90.00] | [−74.47, 153.00] |
| J5 | [−117.60, 117.60] | [−109.92, 57.03] | [−100.00, 116.19] |
| J6 | [−180.00, 180.00] | [−90.00, 80.03] | [−68.79, 139.18] |

Ranges are extrema of recorded joint coordinates in each split. All configurations are observed at the same camera station.

## Supplementary Table 2 · Robot-specific representation evaluation

### 2a · JAKA held-out rendering

| Group | n | PSNR (dB) | SSIM | LPIPS | IoU | Centroid (px) |
|---|---|---|---|---|---|---|
| All | 31 | 30.996 | 0.98250 | 0.01829 | 0.85832 | 3.826 |
| Single-joint | 10 | 32.315 | 0.98569 | 0.01718 | 0.89342 | 3.588 |
| Multi-joint | 21 | 30.368 | 0.98098 | 0.01882 | 0.84160 | 3.939 |

Values are arithmetic means over images for the final rendering model. All 31 test images are included; single-joint and multi-joint subsets are reported separately.

### 2b · Laikago assignment correction

Both versions share reference Gaussians and metric axes. They retain or skip two rounds of reprojection and neighbourhood correction, then both undergo multi-joint refinement. Evaluation uses the same 213 images before hip merging and footprint pruning.

| Assignment correction | PSNR (dB) | IoU | Extra foreground pixels |
|---|---|---|---|
| Retained | 26.945 | 0.9334 | 513.1 |
| Skipped | 25.737 | 0.9152 | 684.3 |

## Supplementary Table 3 · Composition and use of gripper opening

### 3a · Opening and arm-configuration combinations

A separate set of 4,000 eight-DoF configurations is sorted by opening into five groups of 800. Setting A uses the recorded opening and setting B sets it to zero. Both use the end-link projection box under A with a 16-pixel margin.

| Opening group | 1 | 2 | 3 | 4 | 5 | All |
|---|---|---|---|---|---|---|
| Images | 800 | 800 | 800 | 800 | 800 | 4,000 |
| Regional PSNR gain A−B (dB) | 0.008 | 0.080 | 0.167 | 0.209 | 0.240 | 0.141 |

Regional PSNR is higher under A in 3,127 frames. Mean full-image gain is 0.069 dB. Per-finger displacement is sampled continuously and uniformly over 0–0.04 m. All 4,000 values are distinct and cover 401 grid values when quantized at 0.1 mm.

### 3b · Matching object size and opening

| Setting | Geometry criteria met | Conditions |
|---|---|---|
| Adaptive opening | 108/108 | 36 radii × 3 positions |
| Fixed opening | 0/108 | Same conditions, 16 mm per finger |

Twelve sets each contain three radii and three positions. The 36 realized radii span 5.1–34.3 mm, sampled over 5–35 mm. Criteria are midpoint error at most 2 mm, approach-angle error at most 2°, pad-gap error at most 1 mm, and per-finger travel in 0–0.04 m. Maximum adaptive midpoint and approach-angle errors are 0.26 mm and 0.04°. The fitted prismatic direction differs from the simulator reference by approximately 2.0°. The pad gap subtracts closed-state centroid separation.

## Supplementary Table 4 · Metric identification, observation requirements and analytic checks

### 4a · Final-model axis errors

| Model | Direction (°) | Distance (mm) | Distance definition |
|---|---|---|---|
| Panda, image-supervised | 0.459 | 7.477 | Shortest inter-axis distance |
| Panda, metric-anchored | 0.045 | 0.419 | Shortest inter-axis distance |
| Laikago | 0.097 | 0.505 | Shortest inter-axis distance |

Panda averages seven axes and Laikago twelve. Laikago identification observations include 1 mm position noise and 0.25° orientation noise. Axis signs are set by agreement between relative rotation and encoder-angle changes, and final axis errors use shortest inter-axis distance. Laikago's analytic Jacobian agrees with central finite differences over 20 random configurations on all four legs, with maximum absolute difference 1.091 × 10^-10. In-model round-trip checks over 50 targets per leg have maximum residual 9.73 × 10^-4 mm.

The following joint-calibration analyses fit seven Panda axes and two fixed transforms from simulator end-effector observations. They are separate from per-joint identification. Six poses is the necessary parameter-count bound.

### 4b · Observation count

| M | 4 | 8 | 16 | 32 | 64 | 100 |
|---|---|---|---|---|---|---|
| Median held-out error (mm) | 41.998 | 0.0129 | 0.0024 | 0.0099 | 0.0103 | 0.0009 |

| M | 8 | 12 | 16 |
|---|---|---|---|
| Mean held-out error over 10 resamples (mm) | 0.0150 | 0.0034 | 0.0019 |

The first table uses nested subsets. The second independently resamples ten times at each count.

### 4c · Simulated measurement noise

Thirteen conditions comprise a noise-free baseline, four position-noise levels, four orientation-noise levels and four combined levels. Each is repeated ten times at four observation counts, totalling 520 runs. Combined-noise results at $M = 16$ are shown.

| Position and orientation noise | Held-out error (mm) |
|---|---|
| None | 0.0019 |
| 0.5 mm, 0.1° | 1.207 |
| 1 mm, 0.25° | 2.749 |
| 2 mm, 0.5° | 6.061 |
| 5 mm, 1° | 12.240 |

#### 4d · Parameter groups

| Parameters allowed to vary | Held-out error (mm) |
|---|---|
| Fixed transforms only | 27.38 |
| Add axis points | 6.41 |
| Add axis directions | 0.00091 |
| Add joint zero offsets | 0.0282 |

Groups share calibration observations and held-out configurations. Zero offsets couple with axes and fixed transforms, and are fixed in the principal joint-calibration analysis.

### Supplementary Table 5 · Model and solver effects in Panda control

#### 5a · Reaching at fixed targets

| Axes | Solver | Mean (mm) | Median | P95 | Max | Within 5 / 10 mm |
|---|---|---|---|---|---|---|
| Metric-anchored | DLS | 2.280 | 2.26 | 2.60 | 2.75 | 200 / 200 |
| Image-supervised | DLS | 7.636 | 7.37 | 11.36 | 12.55 | 29 / 157 |
| Metric-anchored | Adam | 3.161 | 2.93 | 5.28 | 7.03 | 182 / 200 |
| Image-supervised | Adam | 30.708 | 30.95 | 46.07 | 51.13 | 0 / 1 |

All conditions use the same 200 world-coordinate targets, reference end-effector point and criteria. Errors are measured after simulation execution and all error columns are in millimetres. Targets are sampled with seed 0 from $x \in [0.35, 0.60]$, $y \in [-0.25, 0.25]$, $z \in [0.25, 0.60]$ m. Both solvers use eight starts. Count denominators are 200.

The 95% confidence intervals for mean DLS error are 2.253–2.305 mm and 7.309–7.955 mm for the metric-anchored and image-supervised models. Mean paired difference is 5.357 mm (5.032–5.673 mm), with the image-supervised model worse at every target. DLS mean in-model residuals are about 0.03 mm for both models. Adam in-model residuals are 2.367 mm and 2.326 mm.

#### 5b · Release compensation and model error

| Axes | Offset | Trials | Grasped | Placed | Mean grasp (mm) | Mean place (mm) |
|---|---|---|---|---|---|---|
| Metric-anchored | On | 50 | 50/50 | 50/50 | 2.78 | 3.21 |
| Image-supervised | On | 50 | 50/50 | 50/50 | 13.18 | 19.16 |
| Metric-anchored | Off | 50 | 50/50 | 50/50 | 2.78 | 17.79 |
| Image-supervised | Off | 50 | 50/50 | 49/50 | 13.18 | 31.69 |

Each model has 50 trials per compensation setting. The planar offset $(-2.8, +17.3)$ mm is estimated from separate seed-0 trials, and evaluation uses seed 1. Models are paired by starts, goals, initial configurations and solver settings. Ball radius is 20 mm and planar placement tolerance 40 mm. Mean placement errors include every trial. With compensation, the 95% confidence intervals for mean grasp error are 2.71–2.84 mm and 12.76–13.60 mm. Mean paired difference is 10.40 mm (10.03–10.78 mm).

### Supplementary Table 6 · Laikago branch control and contact conditions

#### 6a · Static foot reaching

| Leg | Targets | Median world error (mm) | Within 5 mm |
|---|---|---|---|
| Front right | 50 | 1.804 | 50/50 |
| Front left | 50 | 0.561 | 50/50 |
| Rear right | 50 | 1.307 | 50/50 |
| Rear left | 50 | 1.072 | 50/50 |
| All | 200 | 1.162 | 200/200 |

The body is fixed and gravity disabled. Targets are sampled without replacement per leg, and solved configurations are directly set in the simulator. Mean world error is 1.256 mm, P95 is 2.725 mm and maximum is 4.122 mm. Median in-model residual is 0.000334 mm. A separate forward check gives median error 1.157 mm.

#### 6b · Friction and torque-limit sensitivity

Friction coefficients $0.6, 0.8, 1.0, 1.2$ and torque limits $40, 100, 200, 500$ N·m form 16 conditions, each evaluated once for 400 control steps. Ranges are minima to maxima across the 16 conditions.

| Quantity | Metric-anchored range | Image-supervised range | Paired direction |
|---|---|---|---|
| Median foot error during walking (mm) | 5.171–8.805 | 39.721–88.258 | Metric-anchored lower in 16/16 |
| Forward displacement (mm) | 244.3–253.2 | 141.0–218.9 | Metric-anchored farther in 16/16 |
| Maximum tilt (°) | 1.04–1.12 | 11.45–11.88 | Metric-anchored lower in 16/16 |

Neither model falls. Figure 6d gives foot-error ratios for every condition.

## Supplementary Table 7 · JAKA prediction and real reaching

| Measure | Self-model | Nominal baseline |
|---|---|---|
| Median FK error, 15 targets (mm) | 6.4 | 11.7 |
| Open loop: executable / median on common 12 | 14/15 / 6.0 mm | 12/15 / 4.8 mm |
| Feedback: valid runs / median initial -> final | 14 / 6.2 -> 4.3 mm | Not evaluated |
| Feedback executions improved | 12/14 | Not evaluated |
| Final error within 5 / 10 mm | 9/14 / 14/14 | Not evaluated |

FK uses all 15 targets; the self-model has lower FK error at 13 of 15. Open-loop medians are paired on the 12 targets executable by both models. Feedback is evaluated on 14 valid self-model executions; final error is the median of three post-stop measurement windows. Target-level records retain failures and measurements.